\documentclass[12pt, twoside]{article}

\usepackage{jmlda}

\newcommand{\hdir}{.}

\newcommand{\bigO}{\mathcal{O}}
\newcommand{\GP}{\mathcal{GP}}
\newcommand{\KL}[2]{\mbox{KL}\left(#1\mbox{$\|$}#2\right)}

\begin{document}
\English

\title
	[Faster variational GP-classification] % short title for page headings, not necessary if a full title fits the headings
    {Faster variational inducing input Gaussian process classification} % full title
\author
	% [D.\,A.~Kropotov, P.\,A.~Izmailov]% short list of the authors (<= 3) for page headings, is necessary only if the full list does not fit the headings
	{P.\,Izmailov$^1$, D.\,Kropotov$^1$, } % full list of the authors, presented in the table of contetns of the issue
    % [F.\,S.~Author$^1$, F.\,S.~Co-Author$^2$, and F.\,S.~Name$^{1, 2}$] % list of the authors presented in the title page of the article, is necessary only if it differs from the full list of the authors in braces, i.e. '{' and '}'
\email
    {izmailovpavel@gmail.com, dmitry.kropotov@gmail.com}
% \thanks
    % {The research was
    %  %partially
    %  supported by the Russian Foundation for Basic Research (grants 00-00-0000 and 00-00-00001).}
\organization
    {$^1$Lomonosov Moscow State University, 1 Leninskie Gory, Moscow, Russia}
\abstract
    {\textbf{Background}: Gaussian processes (GP) provide an elegant and effective approach to learning in kernel machines. This approach leads to a highly interpretable model and allows using the bayesian framework for model adaptation and incorporating the prior knowledge about the problem. GP framework is successfully applied to regression, classification and dimensionality reduction problems. Unfortunately, the standard methods for both GP-regression and GP-classification scale as $\bigO(n^3)$, where $n$ is the size of the dataset, which makes them inapplicable to big data problems. A variety of methods have been proposed to overcome this limitation both for regression and classification problems. The most successful recent methods are based on the concept of inducing inputs. These methods reduce the computational complexity to $\bigO(nm^2)$, where $m$ is the number of inducing inputs with $m$ typically much less, than $n$. In this work we focus on classification. The current state-of-the-art method for this problem is based on stochastic optimization of an evidence lower bound, that depends on $\bigO(m^2)$ parameters. For complex problems, the required number of inducing points $m$ is fairly big, making the optimization in this method challenging.

	\noindent
	\textbf{Methods}: We analyze the structure of variational lower bound that appears in inducing input GP classification. First we notice that using quadratic approximation of several terms in this bound, it is possible to obtain analytical expressions for optimal values of most of the optimization parameters, thus sufficiently reducing the dimension of optimization space. Then we provide two methods for constructing necessary quadratic approximations. One is based on Jaakkola-Jordan bound for logistic function and the other one is derived using Taylor expansion.
    
	\noindent
	\textbf{Results}: We propose two new variational lower bounds for inducing input GP classification that depend on a number of parameters. Then we propose several methods for optimization of these bounds and compare the resulting algorithms with the state-of-the-art approach based on stochastic optimization. Experiments on a bunch of classification datasets show that the new methods perform as well or better than the existing one. However, new methods don't require any tunable parameters and can work in settings within a big range of $n$ and $m$ values thus significantly simplifying training of GP classification models.
	
	}

\maketitle
%%\linenumbers

\typeSection{Introduction}
\noindent %this command is placed at the beginning of the first sentence of each paragraph/section only.
Gaussian processes (GP) provide a prior over functions and allow finding complex regularities in data. Gaussian processes are successfully used for classification/regression problems and dimensionality reduction~\cite{GPinML}. In this work we consider the classification problem only.

Standard methods for GP-classification scale as $\bigO(n^3)$, where $n$ is the size of a training dataset. This complexity makes them inapplicable to big data problems. Therefore, a variety of methods were introduced to overcome this limitations~\cite{sparseGP1,sparseGP2,sparseGP3}. In the paper we focus on methods based on so called inducing inputs. The paper~\cite{Titsias} introduces the inducing inputs approach for training GP models for regression. This approach is based on variational inference and proposes a particular lower bound for marginal likelihood (evidence). This bound is then maximized w.r.t. parameters of kernel function of the Gaussian process, thus fitting the model to data. The computational complexity of this method is $\bigO(nm^2)$, where $m$ is the number of inducing inputs used by the model and is assumed to be substantially smaller than $n$. The paper~\cite{BigData} develops these ideas by showing how to apply stochastic optimization to the evidence lower bound similar to the one used in~\cite{Titsias}. However, a new lower bound depends on $\bigO(m^2)$ variational parameters that makes optimization in the case of big $m$ challenging.

The paper~\cite{SviClass} shows how to apply the approach from~\cite{BigData} to the GP-classification problem. It provides a lower bound that can be optimized w.r.t. kernel parameters and variational parameters using stochastic optimization. However, the lower bound derived in~\cite{SviClass} is intractable and has to be approximated via Gauss-Hermite quadratures or other integral approximation techniques. This lower bound is also fit for stochastic optimization and depends on $\bigO(m^2)$ parameters.

In this work we develop a new approach for training inducing input GP models for classification problems. Here we analyze a structure of variational lower bound from~\cite{SviClass}. We notice that using quadratic approximation of several terms in this bounds it is possible to obtain analytical expressions for optimal values of the most of optimization parameters thus sufficiently reducing the dimension of optimization space. So, we provide two methods for constructing necessary quadratic approximations~--- one is based on Jaakkola-Jordan bound for logistic function, and the other one is derived using Taylor expansion.

The paper is organized as follows. In section~$2$ we describe the standard GP-classification framework and its main limitations. In section~$3$ we introduce the concept of inducing inputs and derive the evidence lower bound of~\cite{SviClass}. Section~$4$ consists of our main contribution - two new tractable evidence lower bounds and different methods for their optimization. Section~$5$ provides experimental comparison of our new methods with the existing approach from~\cite{SviClass} and the last section concludes the paper.

\typeSection{GP-classification model}
%\label{reg_example}
	\label{class_model}
	% \begin{figure}[!t]
% 	\center\includegraphics{\hdir/pictures/1d_gp}
% \caption{One-dimensional Gaussian processes}
% \label{reg_example}
% \end{figure}

\noindent
In this section we review classic Gaussian Processes (GP) framework and its application for classification problems (see~\cite{GPinML} for detailed discussion).

\paragraph{Gaussian process definition}
	\noindent
	A Gaussian process is a collection of random variables, any finite number of which have a joint Gaussian distribution.

	We will only consider processes, that take place in a finite-dimensional real space $\RR^d$. In this case, $f$ is a Gaussian process, if for any $k$, for any $\vec t_1, \dots, \vec t_k \in \RR^d$ the joint distribution
	\[
		(f(\vec t_1), \dots, f(\vec t_k))\T \sim \Normal(\vec m_t, \vec K_t)
	\]
	for some $\vec m_t \in \RR^k$ and $\vec K_t \in \RR^{k \times k}$.

	The mean $\vec m_t$ of this distribution is defined by the mean function $m: \RR^d \rightarrow \RR$ of the Gaussian process:
	\[
		\vec m_t = (m(\vec t_1), \dots, m(\vec t_k))\T.
	\]

	Similarly, the covariance matrix $\vec K_t$ is defined by the covariance function $k:~\RR^d~\times~\RR^d~\rightarrow~\RR$:
	\begin{equation}
        \label{eq:cov_matrix}
		\vec K_t =
		\left (\begin{array}{cccc}
		k(\vec t_1, \vec t_1) & k(\vec t_1, \vec t_2) & \dots & k(\vec t_1, \vec t_n) \\
		k(\vec t_2, \vec t_1) & k(\vec t_2, \vec t_2) & \dots & k(\vec t_2, \vec t_n) \\
		\dots & \dots & \dots & \dots \\
		k(\vec t_n, \vec t_1) & k(\vec t_n, \vec t_2) & \dots & k(\vec t_n, \vec t_n) \\
		\end{array} \right).
	\end{equation}

	It's straightforward then, that a Gaussian process is completely defined by its mean and covariance functions. We will use the following notation:
	\[
		f \sim \GP(m(\cdot), k(\cdot, \cdot))
	\]
	While the mean function $m$ can be an arbitrary real-valued function, the covariance function $k$ has to be a kernel, so that the covariance matrices~\eqref{eq:cov_matrix} it implies are symmetric and positive definite.
	\begin{figure}[!t]
    \begin{center}
    \includegraphics[width=8cm]{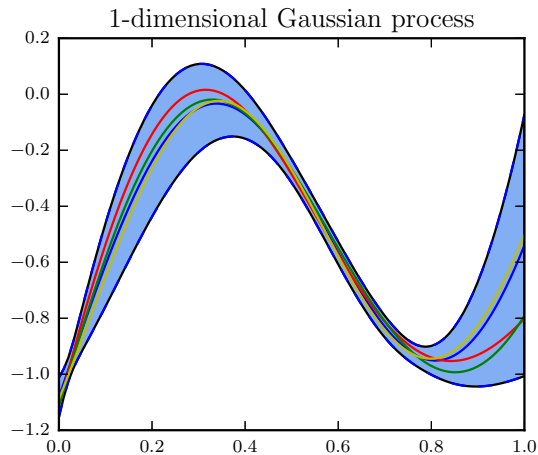}
%    \tabcolsep = 10pt
%        \begin{tabular}{cc}
%            \includegraphics[width=8cm]{\hdir/pictures/1d_gp_samples} & \includegraphics[width=8cm]{\hdir/pictures/1d_gp_ind_inputs}\\
%            (a) & (b)
%        \end{tabular}
    \end{center}
	\caption{One-dimensional Gaussian processes}\label{reg_example}
	\end{figure}

	Fig.~\ref{reg_example} shows an example of a one-dimensional Gaussian process. The dark blue line is the mean function of the process, the light blue region is the $3\sigma$-region, and different color curves are samples from the process.

\paragraph{Gaussian process classification}
	\noindent
	Now we apply Gaussian processes to a binary classification problem. Suppose, we have a dataset $\{(\vec x_i, y_i) \cond i = 1, \dots, n\}$, where $\vec x_i \in \RR^d$, $y_i \in \{-1, 1\}$. Denote the matrix comprised of points $\vec x_1, \dots, \vec x_n$ by $\vec X \in \RR^{n \times d}$ and the vector of corresponding class labels $y_1, \dots, y_n$ by $\vec y \in \{-1, 1\}^n$. The task is to predict the class label $ y_* \in \{-1, 1\}$ at a new point $\vec x_* \in \RR^d$.

	We consider the following model. First we introduce a latent function $f: \RR^d \rightarrow \RR$ and put a zero-mean GP prior over it:
	\[
		f \sim \GP(0, k(\cdot, \cdot))
	\]
	for some covariance function $k(\cdot, \cdot)$. For now the covariance function is supposed to be fixed.

	Then we consider the probability of the object $\vec x_*$ belonging to positive class to be equal to $\sigma(f(\vec x_*))$ for the chosen sigmoid function $\sigma$:
	\begin{equation}
        \label{eq:logistic_likelihood}
		p(y_* = +1 \cond \vec x_*) = \sigma(f(\vec x_*)).
	\end{equation}
	In this work we use the logistic function $\sigma(z) = (1 + \exp(-z))^{-1}$, however one could use other sigmoid functions as well.

	\begin{figure}[!t]
			\center\includegraphics{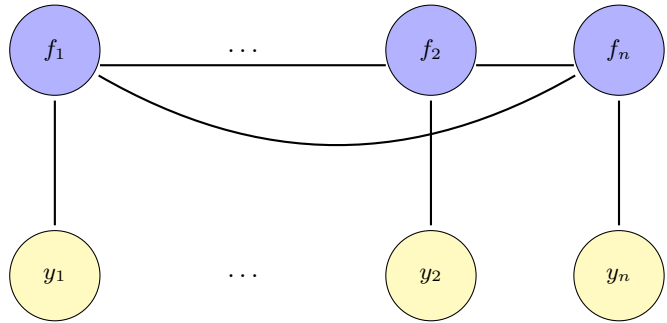}
		\caption{GP-classification graphical model}
		\label{gpc_gm}
	\end{figure}
	\noindent

	The probabilistic model for this setting is given by
	\begin{equation}
        \label{eq:GP_prob_model}
		p(\vec y, \vec f \cond \vec X) = p(\vec y \cond \vec f) p(\vec f \cond \vec X) = p(\vec f \cond \vec X)\prod_{i=1}^np(y_i \cond f_i),
	\end{equation}
	where $p(y_i \cond f_i)$ is sigmoid likelihood~\eqref{eq:logistic_likelihood} and $p(\vec{f} \cond \vec{X})=\Normal(\vec{f} \cond \vec{0},K(\vec X,\vec X))$ is GP prior. The corresponding probabilistic graphical model is given in fig.~\ref{gpc_gm}.
	
    Now inference in the model~\eqref{eq:GP_prob_model} can be done in two steps. First, for new data point $\vec x_*$ we should find the conditional distribution of the corresponding value of the latent process $f_*$. This can be done as follows
	\begin{equation}
		\label{classification_conditional}
		p(f_* \cond \vec y, \vec X, \vec x_*) = \int p(f_* \cond \vec f, \vec X, \vec x_*) p(\vec f \cond \vec y, \vec X) d\vec f.
	\end{equation}
	Second, the probability that $\vec x_*$ belongs to the positive class is obtained by marginalizing over the latent variable $f_*$.
	\begin{equation}
		\label{classification_class_probability}
		p(y_* = +1 \cond \vec y, \vec X, \vec x_*) = \int \sigma(f_*) p(f_* \cond \vec y, \vec X, \vec x_*) df_*.
	\end{equation}

	Unfortunately, both integrals~\eqref{classification_conditional} and~\eqref{classification_class_probability} are intractable since they involve a product of sigmoid functions and normal distributions. Thus, we have to use some integral-approximation techniques to estimate the predictive distribution.

	For example, one can use Laplace approximation method, which builds a Gaussian approximation $q(\vec f \cond \vec y, \vec X)$ to the true posterior $p(\vec f \cond \vec y, \vec X)$. Substituting this Gaussian approximation back into~\eqref{classification_conditional} we obtain a tractable integral. The predictive distribution~\eqref{classification_class_probability} remains intractable, but since this is a one-dimensional integral it can be easily estimated by quadratures or other techniques. The more detailed derivation of this algorithm and another algorithm, based on Expectation Propagation, can be found in~\cite{GPinML}.

	Computational complexity of computing the predictive distribution both for the Laplace approximation method and Expectation propagation scales as $\bigO(n^3)$ since they both require to invert $n{\times}n$ matrix $K(\vec X, \vec X)$. In section~$3$ we describe the concept of inducing points aimed to reduce this complexity.

\paragraph{Model adaptation}
	\label{model_adapt}
	\noindent
	In the previous section we described how to fit a Gaussian process to the data in the classification problem. However, we only considered Gaussian processes with fixed covariance functions. This model can be rather limiting.

	Most of the popular covariance functions have a set of parameters, which we refer to as covariance (or kernel) hyper-parameters. For example, the squared exponential covariance function
	\[
		k_{SE}(\vec x, \vec x'; \vec \theta) = \sigma^2 \exp\left( - \frac{\|\vec x - \vec x'\|^2}{l^2}\right)
	\]
	has two parameters $\vec \theta$~-- variance $\sigma$ and length-scale $l$. An example of a more complicated popular covariance function is the Matern function, given by
	\[
		k_{Matern}(\vec x, \vec x'; \vec \theta) = \frac{2^{1 - \nu}} {\Gamma(\nu)} \left(\frac{\sqrt{2 \nu}\|\vec x - \vec x'\|}{l}\right)^{\nu} K_{\nu} \left( \frac{\sqrt{2 \nu}}{\|\vec x - \vec x'\|}{l}\right),
	\]
	with two positive parameters $\vec \theta = (\nu, l)$. Here $K_{\nu}$ is a modified Bessel function.

	In order to get a good model for the data, one should find a good set of kernel hyper-parameters~$\vec \theta$. Bayesian paradigm provides a way of tuning the kernel hyper-parameters of the GP-model through maximization of the model evidence (marginal likelihood), that is given by
	\begin{equation}
		p(\vec y \cond \vec X, \vec \theta) = \int p(\vec y \cond \vec f) p(\vec f \cond \vec X, \vec \theta) d\vec f \rightarrow \max_{\vec \theta}.
        \label{eq:max_evidence}
	\end{equation}
	However, this integral is intractable for the model~\eqref{eq:GP_prob_model} since it involves a product of sigmoid functions and normal distribution. In subsequent sections we describe several methods to construct a variational lower bound to the marginal likelihood. Maximizing this lower bound with respect to kernel hyper-parameters $\vec \theta$, one could fit the model to the data.

        %\value{sectionCounter}
\typeSection{Variational inducing point GP-classification}
	\label{var_class}
	\noindent
In the previous section we showed how Gaussian processes can be applied to solve classification problems. The computational complexity of GP for classification scales as $\bigO(n^3)$, that makes this method inapplicable to big data problems.

A number of approximate methods have been proposed in the literature for both GP-regression and GP-classification~\cite{sparseGP1,sparseGP2,sparseGP3}. In this paper we consider methods based on the concept of inducing inputs. These methods construct an approximation based on the values of the process at some $m < n$ points. These points are referred to as inducing points. The idea is the following. The hidden Gaussian process $f$ corresponds to some smooth low-dimensional surface in $\RR^d$. This surface can in fact be well approximated by another Gaussian process with properly chosen $m$ training points $\vec Z = (\vec{z}_1,\dots,\vec{z}_m)\T \in \RR^{m{\times}d}$ and process values at that points $\vec{u} = (u_1,\dots,u_m)^T$ (inducing inputs). Then predictions of this new process at training points are used for constructing approximate posterior distribution for $p(\vec f \cond \vec{y}, \vec X)$.
%Fig.~\ref{???} shows an example of one-dimensional Gaussian process and its approximate GP with inducing inputs. 
The positions $\vec Z$ of inducing inputs can be learned within training procedure. However, for simplicity in the following we clusterize the dataset $\vec X$ into $m$ clusters using K-means and choose $\vec Z$ to be cluster centres. In practice we observe that this approach works well in almost all the cases.

%The first methods of this kind chose the inducing points from the training set heuristically or through greedy optimization of some criterion. We will consider the variational approach to selecting the inducing variables. In this approach the inducing inputs are not limited to belong to the training set and their positions as well as the process values at these points can be learned jointly with the values of kernel hyper-parameters. In these variational methods a lower bound for the marginal likelihood, that is simpler to compute than the marginal likelihood itself, is maximized with respect to kernel hyper-parameters to fit the model to the data.

\paragraph{Evidence lower bound}
	\label{vi_jj}
	\begin{figure}[!t]
			\center\includegraphics{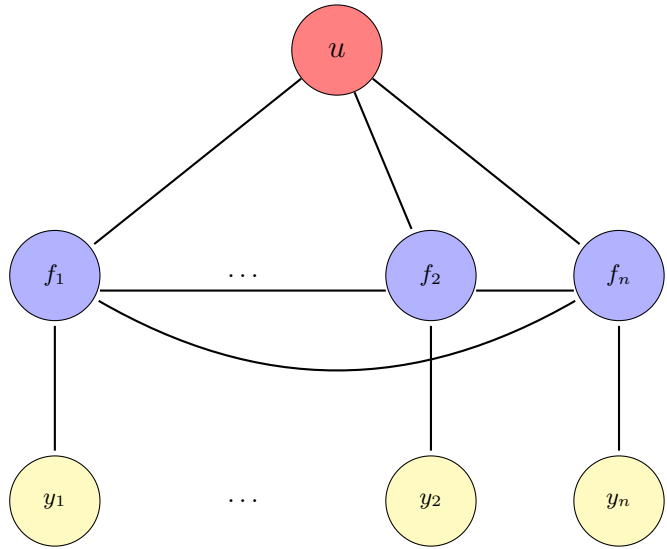}
		\caption{GP-classification graphical model}
		\label{gpc_inducing_gm}
	\end{figure}
	\noindent
    In the following we use a variational approach for solving maximum evidence problem~\eqref{eq:max_evidence}. In this approach an evidence lower bound is introduced, that is simpler to compute than the evidence itself. Then this lower bound is maximized w.r.t. kernel hyperparameters $\vec \theta$ and additional variational parameters used for constructing the lower bound.

	Let's consider the following augmented probabilistic model:
	\begin{equation}\label{augmented_model}
		p(\vec y, \vec f, \vec u \cond \vec X, \vec Z) = p(\vec y \cond \vec f) p(\vec f, \vec{u} \cond \vec X, \vec Z) = \prod_{i = 1}^n p(y_i \cond f_i) p(\vec f ,\vec u \cond \vec X, \vec Z).
	\end{equation}
    %Here $\vec Z = (\vec z_1, \dots, \vec z_m)\T \in \RR^{m \times d}$ are positions of inducing inputs and $\vec{u}$~-- values of the same Gaussian process at points $\vec Z$. 
        The graphical model for the model~\eqref{augmented_model} is shown in fig.~\ref{gpc_inducing_gm}. Note that marginalizing the model~\eqref{augmented_model} w.r.t $\vec u$ gives the initial model~\eqref{eq:GP_prob_model}.

	We denote the covariance matrix comprised of pairwise values of the covariance function $k(\cdot,~\cdot)$ on the points $\vec Z$ by $K(\vec Z, \vec Z) = \vec K_{mm} \in \RR^{m \times m}$. Similarly, we define $\vec K_{nn} = K(\vec X, \vec X) \in \RR^{n \times n}$ and $\vec K_{nm} = K(\vec X, \vec Z) = \vec K_{mn}\T \in \RR^{n \times m}$.

	As $\vec u$ and $\vec f$ are generated from the same Gaussian process with zero-mean prior,
    \begin{align}
	    &p(\vec f, \vec{u}\cond \vec X, \vec Z) = \Normal([\vec f, \vec u] \cond [\vec{0},\vec{0}], K([\vec X, \vec Z], [\vec X, \vec Z])),\notag\\
        &p(\vec u \cond \vec Z) = \Normal(\vec u \cond \vec 0, \vec K_{mm}),\notag\\
		&p(\vec f \cond \vec u, \vec X, \vec Z) = \Normal(\vec f \cond \vec K_{nm} \vec K_{mm}^{-1} \vec u, \vec {\tilde K}),\label{eq:pfu}
	\end{align}
	where $\vec{\tilde K} = \vec K_{nn} - \vec K_{nm} \vec K_{mm}^{-1} \vec K_{mn}$. In the following for simplicity we omit dependence on $\vec X$ and $\vec Z$ in all formulas. Note that we do not consider optimization w.r.t. these values.

	Applying the standard variational lower bound (see, for example \cite{VarBayes}) to the augmented model~\eqref{augmented_model}, we obtain the following inequality:
	\[
		\log p(\vec y) \ge \Expect_{q(\vec u, \vec f)} \log \frac {p(\vec y, \vec u, \vec f)}{q(\vec u, \vec f)} = \Expect_{q(\vec u, \vec f)}\log p(\vec y \cond \vec f) - \KL{q(\vec u, \vec f)} {p(\vec u, \vec f)}
	\]
	for any distribution $q(\vec u, \vec f)$. This inequality becomes equality for the true posterior distribution $q(\vec u, \vec f) = p(\vec u, \vec f \cond \vec y)$. Next we restrict the variational distribution $q(\vec u, \vec f)$ to be of the form
	\begin{equation}
		q(\vec u, \vec f) = p(\vec f \cond \vec u) q(\vec u),
        \label{eq:IP_approx}
	\end{equation}
	where $q(\vec u) = \Normal(\vec u \cond \vec \mu, \vec \Sigma)$ for some $\vec \mu \in \RR^m$, $\vec \Sigma \in \RR^{m \times m}$ and $p(\vec{f}\cond\vec{u})$ is determined by~\eqref{eq:pfu}. This is the key approximation step in inducing points approach for Gaussian processes. The chosen family~\eqref{eq:IP_approx} subsumes that with large enough $m$ all information about the hidden process values $\vec f$ at training points can be successfully restored from the values $\vec u$ at inducing inputs, i.e. $p(\vec f \cond \vec u, \vec y) \approx p(\vec f \cond \vec u)$.

	The form~\eqref{eq:IP_approx} of the variational distribution implies a Gaussian marginal distribution
	\[
		q(\vec f) = \int p(\vec f \cond \vec u) q(\vec u) d\vec u = \Normal(\vec f\cond \vec K_{nm} \vec K_{mm}^{-1} \vec \mu, \vec K_{nn} + \vec K_{nm} \vec K_{mm}^{-1}(\vec \Sigma - \vec K_{mm}) \vec K_{mm}^{-1} \vec K_{mn}).
	\]

	As $\log p(\vec y\cond \vec f)$ depends on $\vec u$ only through $\vec f$, the expectation
	\[
		\Expect_{q(\vec u, \vec f)}\log p(\vec y \cond \vec f) = \Expect_{q(\vec f)}\log p(\vec y \cond \vec f) = \sum_{i = 1}^{n} \Expect_{q(f_i)}\log p(y_i \cond f_i),
	\]where $q(f_i)$ is the marginal distribution of $q(\vec f)$:
	\begin{equation}
		\label{q_f_i}
		q(f_i) = \Normal(f_i \cond \vec k_i\T \vec K_{mm}^{-1} \vec \mu, \vec K_{ii} + \vec k_i^T \vec K_{mm}^{-1} (\vec \Sigma - \vec K_{mm}) \vec K_{mm}^{-1} \vec k_{i}) = \Normal(f_i \cond m_i, S_i^2),
	\end{equation}
	and $\vec k_i$ is the $i$-th column of matrix $\vec K_{mn}$.

	Finally,
	\[
		\KL{q(\vec u, \vec f)} {p(\vec u, \vec f)} = \KL{q(\vec u) p(\vec f \cond \vec u)} {p(\vec u) p(\vec f \cond \vec u)} = \KL{q(\vec u)}{p(\vec u)}.
	\]

	Combining everything back together, we obtain the evidence lower bound
	\begin{equation}
		\label{main_elbo}
		\log p(\vec y) \ge \sum_{i = 1}^{n} \Expect_{q(f_i)} \log p(y_i \cond f_i) - \KL{q(\vec u)} {p(\vec u)}.
	\end{equation}

	Note that the KL-divergence term in the lower bound (\ref{main_elbo}) can be computed analytically since it is a KL-divergence between two normal distributions. In order to compute the expectations $\Expect_{q(f_i)} \log p(y_i \cond f_i)$, we have to use integral approximating techniques.

	The evidence lower bound (ELBO) (\ref{main_elbo}) can be maximized with respect to variational parameters $\vec \mu$, $\vec \Sigma$ and kernel hyper-parameters. Using the optimal distribution $q(\vec u)$, we can perform predictions for new data point $\vec x_*$ as follows
	\begin{multline*}
		p(f_* \cond \vec y) = \int p(f_* \cond \vec u, \vec f) p(\vec u, \vec f\cond \vec y) d\vec u d\vec f \approx \int p(f_* \cond \vec u, \vec f) q(\vec u, \vec f) d\vec u d\vec f = \\
		= \int p(f_* \cond \vec u, \vec f) p(\vec f \cond \vec u) q(\vec u) d\vec u d\vec f = \int p(f_* \cond \vec u) q(\vec u) d\vec u.
	\end{multline*}
	The last integral is tractable since both terms $p(f_* \cond \vec u)$ and $q(\vec u)$ are normal distributions.

        Note, that in case of regression with Gaussian noise the distributions $p(y_i \cond f_i)$ are Gaussians and thus the expectations $\Expect_{q(f_i)} \log p(y_i \cond f_i)$ are tractable. Paper~\cite{BigData} suggests  maximization of the lower bound~\eqref{main_elbo} with respect to $\vec \mu$, $\vec \Sigma$ and covariance hyper-parameters with stochastic optimization techniques for GP-regression.

\paragraph{SVI method}
	\noindent
    In case of classification we can't analytically compute the expectations $\Expect_{q(f_i)} \log p(y_i \cond f_i)$ in the lower bound (\ref{main_elbo}). However, the expectations are one-dimensional Gaussian integrals and can thus be effectively approximated with a range of techniques. In paper~\cite{SviClass} Gauss-Hermite quadratures are used for this purpose. Note that the lower bound~\eqref{main_elbo} has the form ''sum over training objects''. Hence this bound can be maximized using stochastic optimization techniques. Paper~\cite{SviClass} suggests to maximize the lower bound~\eqref{main_elbo} with respect to the variational parameters $\vec \mu, \vec \Sigma$ and kernel hyper-parameters $\vec \theta$ using stochastic optimization. We refer to this method as \verb'svi' method (abbreviation for Stochastic Variational Inference). The lower bound (\ref{main_elbo}) and all it's derivatives can be computed in $\bigO(n m^2 + m^3)$. This complexity has a linear dependence on $n$, hence \verb'svi' method can be applied for the case of big training data.

        %\value{sectionCounter}
\section{Tractable evidence lower bound for GP-classification}
	\label{quadratic_approx}
	\noindent
In the previous section we described the \verb'svi' method. It is based on stochastic optimization of the lower bound (\ref{main_elbo}) for marginal likelihood and the lower bound itself is computed in $\bigO(nm^2)$. But the bound depends on $\bigO(m^2)$ parameters which makes the optimization problem hard to solve when a big number of inducing points is needed.

For GP-regression the situation is similar. Paper~\cite{BigData} describes a method analogical to the \verb'svi' method for classification. The only difference is that the lower bound becomes tractable in case of regression. Then the paper~\cite{Titsias} tries to solve the problem of big $\bigO(m^2)$ number of parameters in the algorithm from~\cite{BigData} in the following way. In case of regression the lower bound~\eqref{main_elbo} can be analytically optimised with respect to variational parameters $\vec \mu, \vec \Sigma$. Doing so and substituting the optimal values back into the lower bound, one can obtain a new lower bound to the marginal likelihood that depends solely on kernel hyper-parameters $\vec \theta$. This simplifies the optimization problem by dramatically reducing the number of optimization parameters. Unfortunately, this new bound doesn't have a form of ''sum over objects'', hence stochastic optimization methods are no longer applicable here. However, in our experiments we've found that even for fairly big datasets the method from~\cite{Titsias} outperforms~\cite{BigData} despite the lack of stochastic optimization.

In the following subsection we devise an approach similar to the method of~\cite{Titsias} for the case of classification. We provide a tractable evidence lower bound and analytically maximize it with respect to variational parameters $\vec \mu$ and $\vec \Sigma$. Substituting the optimal values of these parameters back into the lower bound we obtain a new lower bound, that depends only on kernel hyper-parameters~$\vec \theta$.

\paragraph{Global evidence lower bound}
	\noindent
	In order to derive a tractable lower bound for~\eqref{main_elbo}, we seek a quadratic approximation to the log-logistic function $\log p(y_i \cond f_i) = \log \sigma(y_i f_i)$, where $\log \sigma(t) = - \log(1 + \exp(-t))$. The paper~\cite{JJ} provides a global parametric quadratic lower bound for this function:

	% We will rederive this bound here.

	% \[
	% 	\log g(t) = - \log(1 + \exp(-t)) = \frac t 2 - \log\left(\exp\left(\frac t 2\right) - \exp\left(-\frac t 2\right)\right).
	% \]
	% Now, consider the function $f(t) = \log\left(\exp\left(\frac t 2\right) - \exp\left(-\frac t 2\right) \right)$. Note that $f$ is a convex function in the variable $t^2$. A tangent surface to a convex function is a global lower bound for this function. Performing the first-order Taylor expansion in the variable $t^2$ at the point $\xi_t$ we obtain
	% \[
	% 	f(t) \ge f(\xi_t) + \derivative{f(\xi_t)}{\xi_t} (t^2 - \xi_t^2) = -\frac {\xi_t} 2 + \log g(\xi_t) + \frac 1 {4 \xi_t} \tanh\left(\frac {\xi_t} 2 \right) (t^2 - \xi_t^2),
	% \]
	% for any point $\xi_t$. This bound becomes tight, when $\xi_t = t$. Denoting
	% \[
	% 	\lambda(\xi_t) = \frac {\tanh\left(\frac{\xi_t} 2\right)}{4 \xi_t}.
	% \]
	% we can rewrite
	\[
		\log \sigma(t) \ge \frac t 2 - \frac {\xi_t} 2 + \log \sigma(\xi_t) - \lambda(\xi_t) (t^2 - \xi_t^2),\ \forall t
	\]
	where $\lambda(\xi_t) = \tanh(\xi_t) / (4\xi_t)$ and $\xi_t \in \RR$ is a parameter of the bound. This bound is tight when $t^2 = \xi_t^2$.

	Substituting this bound back to~\eqref{main_elbo} with separate values $\vec \xi = \{\xi_i \cond i = 1, \ldots, n\}$ for every data point, we obtain a tractable lower bound
	\[
		\log p(y) \ge \sum_{i = 1}^{n} \Expect_{q(f_i)} \log p(y_i | f_i) - \KL{q(\vec u)} {p(\vec u)} = \sum_{i = 1}^{n} \Expect_{q(f_i)} \log \sigma(y_i f_i) - \KL{q(\vec u)} {p(\vec u)} \ge
	\]
	\[
		\ge \sum_{i = 1}^{n}\left(\Expect_{q(f_i)} \left [\log \sigma(\xi_i) + \frac {y_i f_i - \xi_i} {2} - \lambda(\xi_i) (f_i^2 - \xi_i^2) \right]\right) - \KL{q(\vec u)} {p(\vec u)} =
	\]
	\[
		= \sum_{i = 1}^{n} \left(\log \sigma(\xi_i) - \frac {\xi_i}{2} + \lambda(\xi_i) \xi_i^2\right) + \frac 1 2 \vec \mu\T \vec K_{mm}^{-1} \vec K_{mn} \vec y - \Tr\left(\vec \Lambda(\vec \xi) (\vec K_{nn} + \vec K_{nm} \vec K_{mm}^{-1} (\vec \Sigma - \vec K_{mm}) \vec K_{mm}^{-1} \vec K_{mn})\right) -
	\]
	\[
		- \vec \mu\T \vec K_{mm}^{-1} \vec K_{mn} \vec \Lambda(\vec \xi) \vec K_{nm} \vec K_{mm}^{-1} \vec \mu - \KL{q(\vec u)} {p(\vec u)} = J(\vec \mu, \vec \Sigma, \vec \xi, \vec \theta),
	\]
	where
	\[
		\vec \Lambda(\vec \xi) =
		\left(
		\begin{array}{cccc}
			\lambda(\xi_1) & 0 & \dots & 0 \\
			0 & \lambda(\xi_2) & \dots & 0 \\
			\ldots & \dots & \dots & \ldots \\
			0 & 0 & \dots & \lambda(\xi_n) \\
		\end{array}
		\right).
	\]

	Differentiating $J$ with respect to $\vec \mu$ and $\vec \Sigma$ and setting the derivatives to zero, we obtain
	\begin{equation}\label{optimal_sigma}
		\vec {\hat \Sigma}(\vec \xi) = (2 \vec K_{mm}^{-1} \vec K_{mn} \vec \Lambda(\vec \xi) \vec K_{nm} \vec K_{mm}^{-1} + \vec K_{mm}^{-1})^{-1},
	\end{equation}
	\begin{equation}\label{optimal_mu}
		\vec {\hat \mu}(\vec \xi) = \frac 1 2 \vec {\hat \Sigma}(\vec \xi) \vec K_{mm}^{-1} \vec K_{mn} \vec y.
	\end{equation}

	Substituting the optimal values of variational parameters back to the lower bound $J$ and omitting the terms not depending on $\vec \theta$ and $\vec \xi$, we obtain a compact lower bound
	\begin{equation}\label{J_xi}
		\begin{split}
		\hat J(\vec \theta, \vec \xi) = \sum_{i=1}^n \left(\log \sigma(\xi_i) - \frac{\xi_i}{2} + \lambda(\xi_i) \xi_i^2\right) + \frac 1 8 \vec y\T \vec K_{nm} \vec B^{-1} \vec K_{mn} \vec y \\ + \frac 1 2 \log |\vec K_{mm}| - \frac 1 2 \log |\vec B| - \Tr(\vec \Lambda(\vec \xi) \vec {\tilde K}),
		\end{split}
	\end{equation}
	where
	\[
		\vec {\tilde K} = \vec K_{nn} - \vec K_{nm} \vec K_{mm}^{-1} \vec K_{mn},
	\]
	\[
		\vec B = 2 \vec K_{mn} \vec \Lambda(\vec \xi) \vec K_{nm} + \vec K_{mm}.
	\]

	In the following we consider three different methods for maximizing the lower bound $\hat J(\vec \theta, \vec \xi)$.

	\begin{algorithm}
		\caption{vi-JJ method}
		\label{vi_jj}
		\begin{algorithmic}

		\REQUIRE $n_{upd}$, $n_{fun}$
		\ENSURE $\vec\theta$, $\vec\mu$, $\vec\Sigma$
		\STATE $\vec \mu$, $\vec\Sigma \leftarrow \vec 0$, $\vec I$
		\REPEAT
		% \FOR{$i \leftarrow 1, \dots, n_{out}$ :}
			$\tilde{\vec\mu}$, $\tilde{\vec\Sigma} \leftarrow \vec\mu$, $\vec\Sigma$
			\FOR{$j \leftarrow 1, \dots, n_{upd}$ :\COMMENT{stage 1: updating $\vec\mu$, $\vec\Sigma$, $\vec\xi$}}
				\STATE $m_t$, $S_t \leftarrow \vec k_t\T \vec K_{mm}^{-1} \tilde{\vec\mu}$, $\vec K_{tt} + \vec k_t^T \vec K_{mm}^{-1} (\tilde{\vec\Sigma} - \vec K_{mm}) \vec K_{mm}^{-1} \vec k_{t}$, \hspace{0.2cm}$t = 1, \dots, n$
				\STATE $\tilde{\xi_t}^2 \leftarrow m_t^2 + S_t^2$, \hspace{0.2cm}$t = 1, \dots, n$
			\ENDFOR
			\STATE $\tilde{\vec\mu}$, $\tilde{\vec\Sigma} \leftarrow \hat{\vec \mu}(\tilde{\vec\xi})$, $ \hat{\vec\Sigma}(\tilde{\vec\xi})$ \COMMENT{see (\ref{optimal_sigma}), (\ref{optimal_mu}).}
			\STATE $\vec \mu$, $\vec\Sigma$, $\vec \xi \leftarrow \tilde{\vec\mu}$, $\tilde{\vec \Sigma}$, $\tilde{\vec \xi}$
			\STATE $\vec \theta$ = minimize($\hat J(\cdot, \vec \xi)$, method='L-BFGS-B', maxfun=$n_{fun}$)\COMMENT{stage 2: updating $\vec\theta$}
		% \ENDFOR
		\UNTIL{convergence}
		\end{algorithmic}
	\end{algorithm}

	Note, that given the values of $\vec \mu$, $\vec \Sigma$, $\vec \theta$, we can maximize $J(\vec \mu, \vec \Sigma, \vec \xi, \vec \theta)$ with respect to $\vec \xi$ analytically. The optimal values for $\vec \xi$ are given by

	\begin{equation}\label{optimal_xi}
		\xi_i^2 = \Expect_{q(\vec f)} f_i^2 = m_i^2 + S_i^2.
	\end{equation}
	The values $m_i$ and $S_i$ were defined in (\ref{q_f_i}). In our first method we use analytical formulas (\ref{optimal_xi}) to recompute the values of $\vec\xi$ and use gradient-based optimization to maximize the bound with respect to $\vec\theta$. The pseudocode is given in Alg. \ref{vi_jj}. We will refer to this method as \verb'vi-JJ', where \verb'JJ' stands for Jaakkola and Jordan, the authors of \cite{JJ}. Note that the computational complexity of one iteration of this method is $\bigO(nm^2)$, the same as for the \verb'svi' method.

	Our second method uses gradient-based optimization to maximize $\hat J$ with respect to both $\vec\theta$ and $\vec\xi$. Note that in this method we don't have to recompute $\vec \mu$ and $\vec \Sigma$ at each iteration which makes the methods iterations empirically faster for big values of $m$. We refer to this method as \verb'vi-JJ-full'.

	Finally, \verb'vi-JJ-hybrid' is a combination of the two methods described above. The general scheme of this method is the same as \verb'vi-JJ'. In the \verb'vi-JJ-hybrid' method we use analytical formulas to recompute $\vec\xi$ as we do in the \verb'vi-JJ' method at stage $1$, but at stage $2$ we use gradient-based optimization with respect to both $\vec\xi$ and $\vec\theta$. The virtues of this method will be described in the experiments section.

	% On the first step we fix $\vec \theta$ and estimate the optimal values of $\vec \mu$, $\vec \Sigma$, and $\vec \xi$. This is done using the formulas (\ref{optimal_sigma}), (\ref{optimal_mu}), (\ref{optimal_xi}). First, we fix the values of $\vec \xi$ and recompute $\vec \mu$, $\vec \Sigma$, using (\ref{optimal_sigma}), (\ref{optimal_mu}). Then, we fix $\vec \mu$ and $\vec \Sigma$ and recompute $\vec \xi$, using (\ref{optimal_xi}). We repeat this procedure several times.

	% On the second step we fix $\vec \xi$, and maximize the lower bound (\ref{J_xi}) with respect to $\vec \theta$ via gradient-based optimization.

\paragraph{Tractable local approximation to the evidence lower bound}
	\noindent
	Another way to obtain a tractable approximation to the lower bound~\eqref{main_elbo} is to use a local quadratic approximation for the log-logistic function $\log p(y_i \cond f_i)$. In this way we perform a second-order Taylor expansion of this function at points $\vec \xi = \{\xi_i \cond i = 1, \dots, n\}$:

	\begin{equation}
		\log p(y_i \cond f_i) \approx - \log(1 + \exp( - y_i \xi_i)) +  \frac{y_i}{1 + \exp(y_i \xi_i)} (f_i - \xi_i) - \frac{y_i^2 \exp(y_i \xi_i)}{2 (1 + \exp(y_i \xi_i))^2} (f_i - \xi_i)^2.
        \label{eq:taylor_approx}
	\end{equation}

	The following derivation is analogical to the derivation in the previous section. Substituting the approximation~\eqref{eq:taylor_approx} into the lower bound~\eqref{main_elbo}, we obtain

	\[\log p(y) \ge \sum_{i = 1}^{n} \Expect_{q(f_i)} \log p(y_i | f_i) - \KL{q(\vec u)} {p(\vec u)} \approx
	\]
	% $$
	% \approx \sum_{i = 1}^{n}\left( - \log(1 + \exp( - y_i \xi_i)) +  \frac{y_i}{1 + \exp(y_i \xi_i)} (m_i - \xi_i)  - \right.$$
	% $$ \left. - \frac{y_i^2 \exp(y_i \xi_i)}{2 (1 + \exp(y_i \xi_i))^2} \left(S_i^2 + (m_i - \xi_i)^2\right)\right) -\frac 1 2 \left (\log \frac {|K_{mm}|} {|\Sigma|} - m + \Tr(K_{mm}^{-1} \Sigma) + \mu^T K_{mm}^{-1} \mu \right) = $$
	\[
		\approx - \sum_{i = 1}^{n} \log(1 + \exp( - y_i \xi_i)) + \varphi(\vec\xi)^T (\vec K_{nm} \vec K_{mm}^{-1} \vec\mu - \vec\xi) -
	\]
	\[
		- \Tr\left(\vec\Psi(\vec\xi)(\vec K_{nn} + \vec K_{nm} \vec K_{mm}^{-1} (\vec\Sigma - \vec K_{mm})\vec K_{mm}^{-1}\vec K_{mn})\right) - (\vec K_{nm} \vec K_{mm}^{-1} \vec \mu - \vec \xi)^T \vec \Psi(\vec \xi) (\vec K_{nm} \vec K_{mm}^{-1} \vec \mu - \vec \xi) -
	\]
	\[
		- \frac 1 2 \left (\log \frac {|\vec K_{mm}|} {|\vec \Sigma|} - m + \Tr(\vec K_{mm}^{-1} \vec \Sigma) + \vec \mu^T \vec K_{mm}^{-1} \vec \mu \right).
	\]
	Here
	$\vec \Psi(\vec \xi)$ is a diagonal matrix
	\[\vec \Psi(\vec \xi) =
		\left(
		\begin{array}{cccc}
			\psi(\xi_1) & 0 & \dots & 0 \\
			0 & \psi(\xi_2) & \dots & 0 \\
			\dots & \dots & \dots & \dots \\
			0 & 0 & \dots & \psi(\xi_n) \\
		\end{array}
		\right),
	\]
	where $\psi(\xi_i) = \cfrac{y_i^2 \exp(y_i \xi_i)}{2 (1 + \exp(y_i \xi_i))^2}$.

	Differentiating the approximate bound with respect to $\vec \mu$, $\vec \Sigma$, and $\vec \xi$, and setting the derivatives to zero, we obtain the following formulas for optimal values of these parameters:

	\begin{equation}
		\label{optimal_sigma_t}
		\vec{\hat\Sigma}(\vec \xi) = \left(2 \vec K_{mm}^{-1}\vec K_{mn} \vec \Psi(\vec \xi) \vec K_{nm} \vec K_{mm}^{-1} + \vec K_{mm}^{-1}\right)^{-1},
	\end{equation}
	\begin{equation}
		\label{optimal_mu_t}
		\vec {\hat\mu}(\vec \xi) = \vec {\hat\Sigma}(\vec \xi) \vec K_{mm}^{-1} \vec K_{mn} \vec v(\vec \xi),
	\end{equation}
	\[
		\xi_i = m_i.
	\]
	Here
	\[
		\vec v(\vec \xi) = \vec \varphi(\vec \xi) + 2 \vec \Psi(\vec \xi) \vec \xi,
	\]
	and $\vec \varphi(\vec \xi)$ is a vector, composed of
	\[
		\vec \varphi(\vec \xi)_i = \frac{y_i}{1 + \exp(y_i \xi_i)}.
	\]

	Substituting the optimal values for $\vec \mu$ and $\vec \Sigma$, given by (\ref{optimal_sigma_t}), (\ref{optimal_mu_t}), back into the approximate bound, and omiting the terms, that do not depend on $\vec \theta$, we obtain the following approximate lower bound:
	\begin{equation}
		\label{vi_t_elbo}
		\tilde J_{\vec\xi} = \frac 1 2 \vec v(\vec \xi)\T \vec K_{nm} \vec B^{-1} \vec K_{mn} \vec v(\vec \xi) + \frac 1 2 \log |\vec K_{mm}| - \frac 1 2 \log |\vec B| - \Tr(\vec \Psi(\vec \xi) \vec {\tilde K}),
	\end{equation}
	where
	\[
		\vec B = 2 \vec K_{mn} \vec \Psi(\vec \xi) \vec K_{nm} + \vec K_{mm}.
	\]

	Note that the lower bound~(\ref{vi_t_elbo}) is not a global lower bound for the log-evidence $\log p(y)$. However, locally we get a good approximation of the evidence lower bound (\ref{main_elbo}).

	For maximizing the approximate lower bound (\ref{vi_t_elbo}) we consider a method, analogical to \verb'vi-JJ'. In order to specify this method we simply substitute the bound $\hat J(\cdot, \vec\xi)$ by $\tilde J_{\vec \xi}$ in the second stage in Alg.~\ref{vi_jj}. We will refer to this method as \verb'vi-Taylor'. The computational complexity of one iteration of this method is once again $\bigO(nm^2)$.

\section{Experiments}
	\label{experiments}
	\noindent
In this section we empirically compare the derived \verb'vi-JJ', \verb'vi-Taylor', \verb'vi-JJ-full' and \verb'vi-JJ-hybrid' methods with \verb'svi'. Below we describe the setting of the experiments and discuss their results.

\paragraph{Experimental setting}
	\begin{figure}[!t]
		\scalebox{0.85}{
		\subfloat{
			\includegraphics{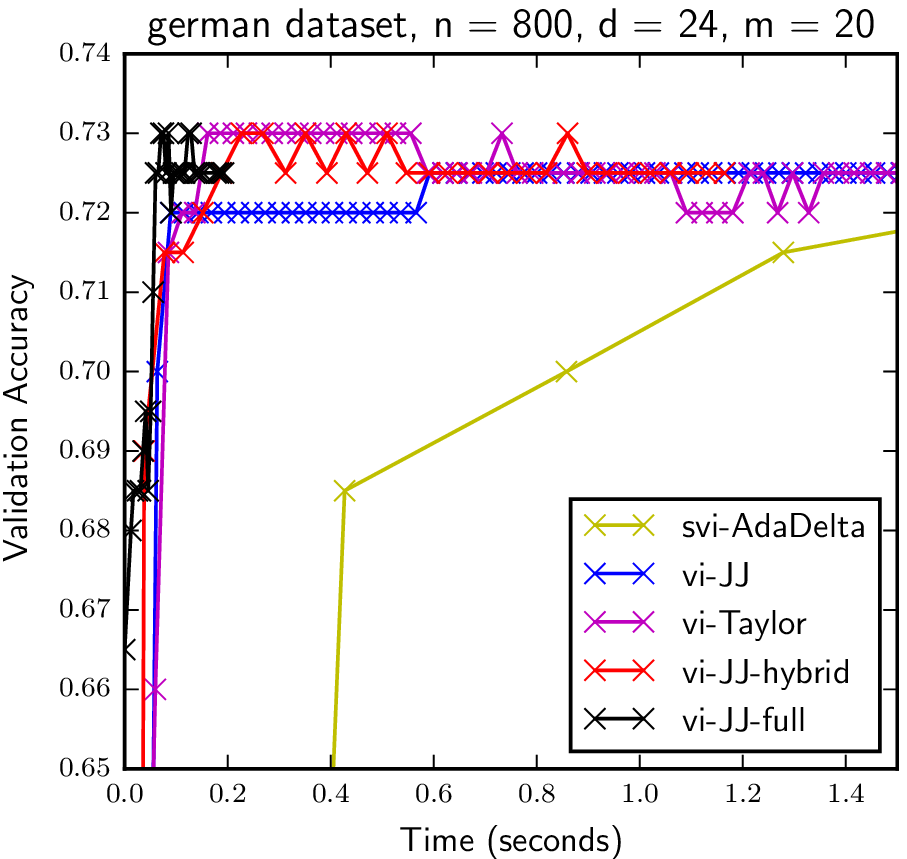}
		}
		\subfloat{
			\includegraphics{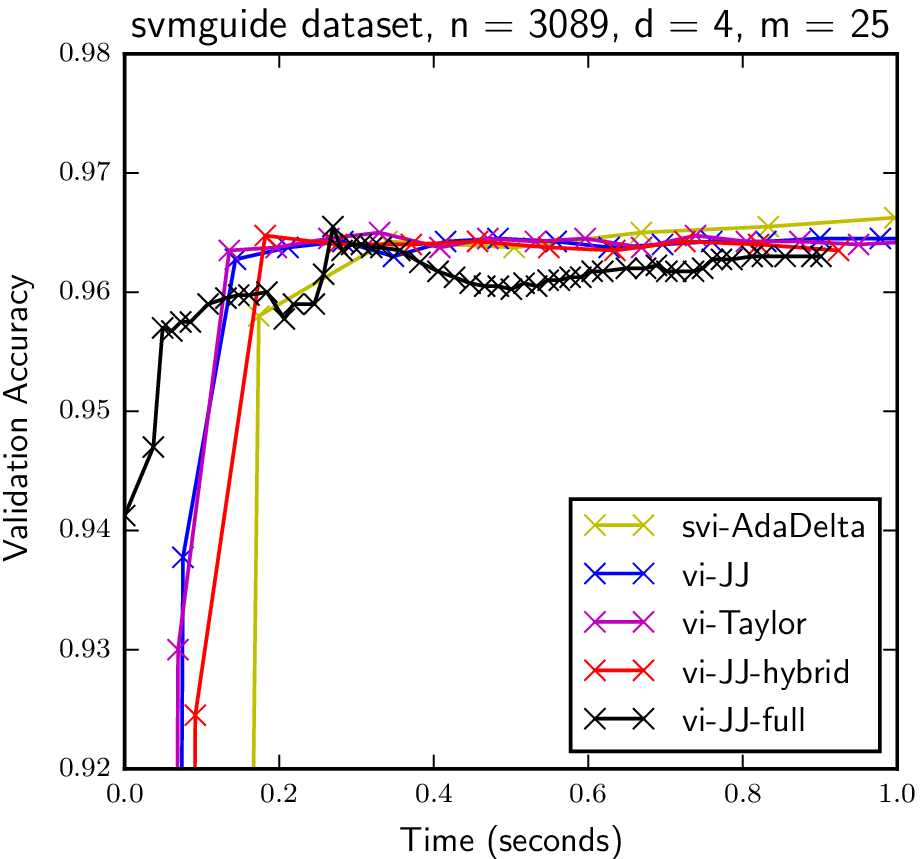}
		}}
		\caption{Methods performance on small datasets}
		\label{small}
	\end{figure}

	\noindent
	In our experiments we compared $5$ methods for variational inducing point GP-classification:
	\begin{itemize}
		\item \verb'svi-AdaDelta' uses the AdaDelta optimization method for maximization of the lower bound (\ref{main_elbo}) as it is done in the paper~\cite{SviClass};
		\item \verb'vi-JJ' was described in section~$4.1$;
		\item \verb'vi-Taylor' was described in section~$4.2$;
		\item \verb'vi-JJ-full' was described in section~$4.1$;
		\item \verb'vi-JJ-hybrid' was described in section~$4.1$.
	\end{itemize}
	We also tried using deterministic L-BFGS-B optimization method for maximizing the evidence lower bound (\ref{main_elbo}), but it worked substantially worse than all the other methods. Note, that all the methods have the same complexity of epochs $\bigO(nm^2)$. Table \ref{methods_table} shows which variables are optimized numerically and which are optimized analytically for each method.

	\begin{table}[b!]
		\center
		\caption{Methods outline}
		\label{methods_table}
		\begin{tabular}{c c c}
			\hline
			Method &
			\begin{tabular}{c} Numerically optimized\\ variables\end{tabular}&
			\begin{tabular}{c} Analytically optimized\\ variables\end{tabular} \\
			\hline
			\verb'svi-AdaDelta' & $\vec \theta$, $\vec \mu \in \RR^m$, $\vec \Sigma \in \RR^{m \times m}$ & \\
			\verb'vi-JJ', \verb'vi-Taylor' & $\vec \theta$ & $\vec \mu \in \RR^m$, $\vec \Sigma \in \RR^{m \times m}$, $\vec \xi \in \RR^n$\\
			\verb'vi-JJ-hybrid' & $\vec \theta$, $\vec \xi \in \RR^n$ & $\vec \mu \in \RR^m$, $\vec \Sigma \in \RR^{m \times m}$, $\vec \xi \in \RR^n$\\
			\verb'vi-JJ-full' & $\vec \theta$, $\vec \xi \in \RR^n$ & $\vec \mu \in \RR^m$, $\vec \Sigma \in \RR^{m \times m}$ \\

			\hline
		\end{tabular}
	\end{table}

	In our experiments we didn't optimize the lower bound with respect to the positions $\vec Z$ of inducing points. Instead we used $K$-means clustering procedure with $K$ equal to the number $m$ of inducing inputs and took clusters centres as $\vec Z$. Also we used the squared exponential covariance function (see section \ref{model_adapt}) in all experiments with a Gaussian noise term.

	The stochastic method \verb'svi-AdaDelta' requires the user to manually specify the learning rate and the batch size for the optimization method. For the former we had to run the method with different learning rates and choose the value that resulted in the fastest convergence. We used the learning rates from a fixed grid with a step of $0.1$. It always happened, that for the largest value from the grid the method diverged, and for the smallest the method converged slower, than for some medium value, verifying, that the optimal learning rate was somewhere in the range. To choose the batch size we used the following convention. For small {\it german} and {\it svmguide} datasets we've set the batch size to $50$. For other datasets we used approximately $\frac n {100}$ as the batch size, where $n$ is the size of the training set.

	\begin{figure}[!t]
		\scalebox{0.85}{
		\subfloat{
			\includegraphics{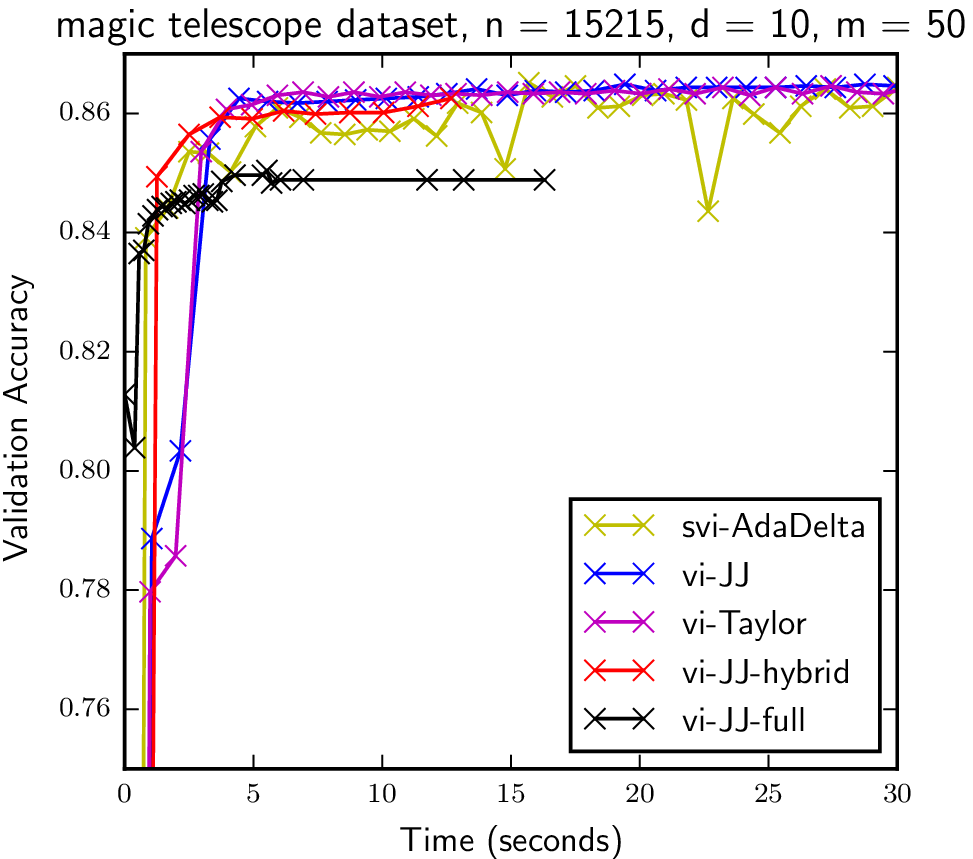}
		}
		\subfloat{
			\includegraphics{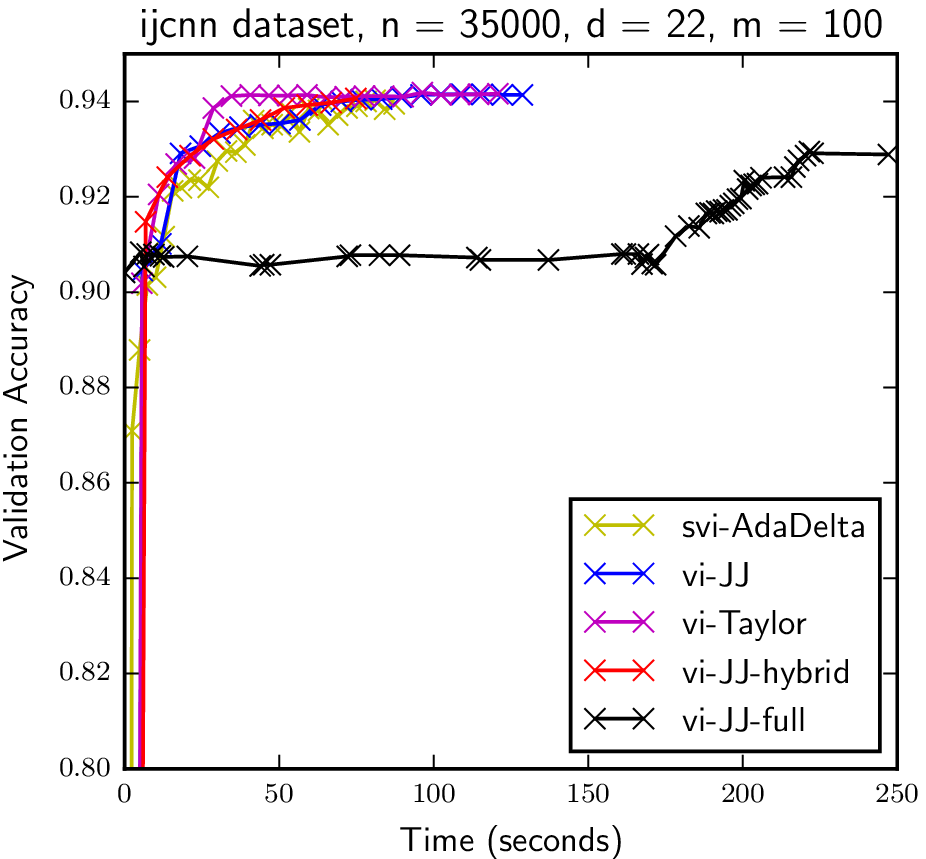}
		}}
		\caption{Methods performance on medium datasets}
		\label{medium}
	\end{figure}
	\noindent

	For the \verb'vi-JJ', \verb'vi-Taylor' and \verb'vi-JJ-hybrid' in all of the experiments on every iteration we recomputed the values of $\vec \xi$, $\vec \mu$, and $\vec \Sigma$ $3$ times ($n_{upd} = 3$ in algorithm \ref{vi_jj}). To tune $\vec \theta$, on every iteration we've run L-BFGS-B optimization method constrained to do no more than $5$ evaluations of the lower bound and it's gradient. We found that these values of parameters work well for all the datasets we experimented with.

	For the \verb'svi-AdaDelta' method we used optimization w.r.t. Cholesky factor of the matrix $\vec \Sigma$ in order to maintain it's positive definiteness, as described in~\cite{SviClass}. We used AdaDelta optimization method implementation from the {\it climin} toolbox~\cite{climin} as is done in the original paper.

	For every dataset we experimented with a number of inducing points to verify that the results of the methods are close to optimal.

	We evaluate the methods plotting the accuracy of their predictions on the test data against time. All of the plots have titles of the following format.
	\[
		\mbox{[name of the dataset]}, n = \mbox{[number of objects in the training set]},
	\]
	\[
		d = \mbox{[number of features]}, m = \mbox{[number of inducing inputs]}
	\]

	We also preprocessed all the datasets by normalizing the features setting the mean of all features to $0$ and the variance to $1$. For datasets without available test data, we used $20\%$ of the data as a test set and $80\%$ as a train set.

	\paragraph{Results and discussion}

		\begin{figure}[!t]
			\scalebox{0.85}{
			\subfloat{
				\includegraphics{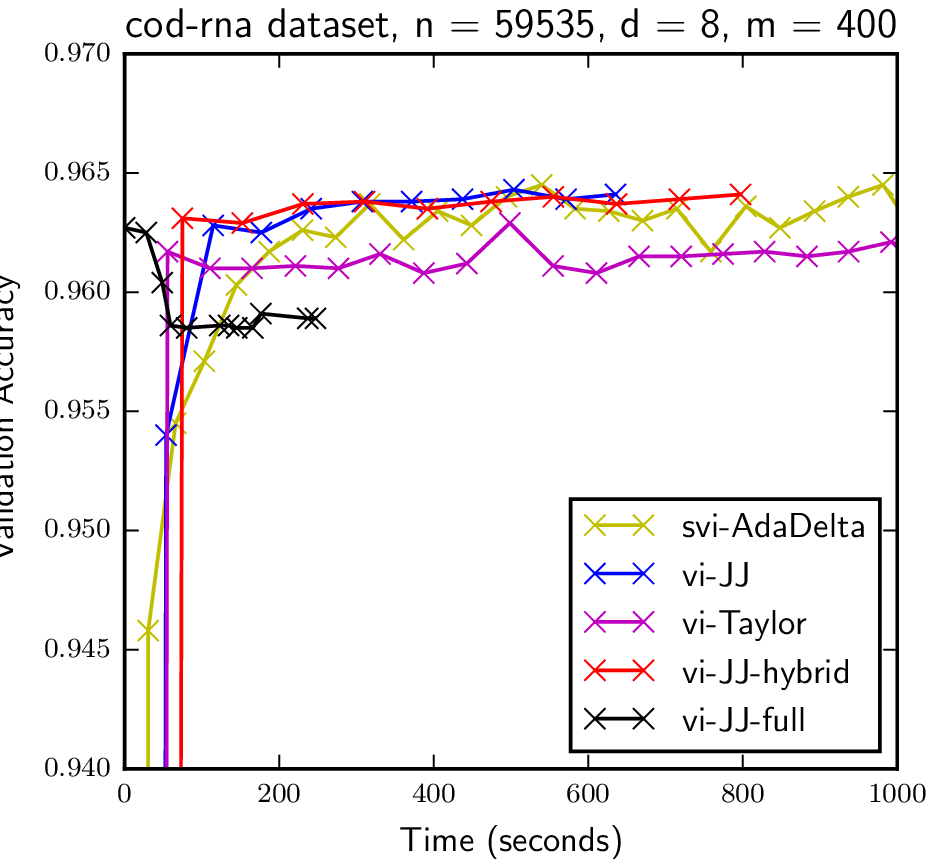}
			}
			\subfloat{
				\includegraphics{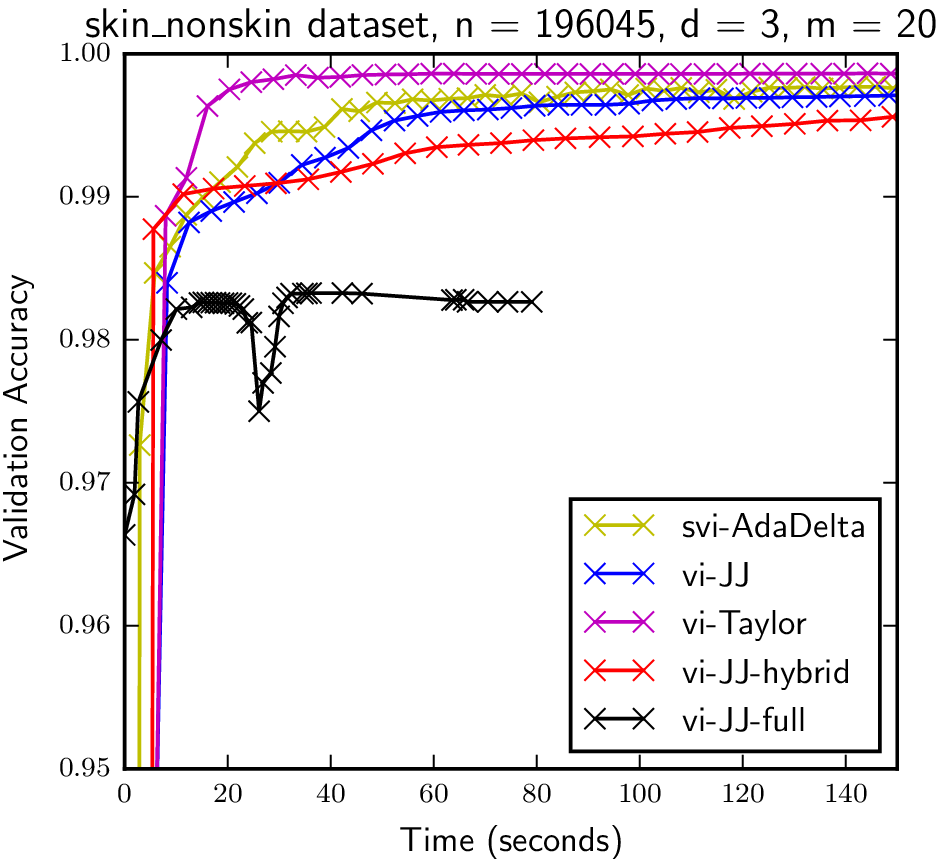}
			}}
			\caption{Methods performance on big datasets}
			\label{big}
		\end{figure}

		\noindent
		We compared the methods' performance on seven datasets. Here we discuss the results.

		Fig. \ref{small} provides the results for {\it german} and {\it svmguide} datasets. As we can see, on the small {\it german} dataset the stochastic \verb'svi-AdaDelta' method struggles, and it takes it longer to achieve the optimal quality, than for all the other methods, which show similar results. On the {\it svmguide} dataset it takes \verb'vi-JJ-full' and \verb'svi-AdaDelta' a little bit longer to converge, while the other three methods show roughly the same performance.

		The results on {\it magic telescope} and {\it ijcnn} datasets are provided in fig. \ref{medium}. On the {\it magic telescope} dataset the \verb'vi-JJ' and \verb'vi-Taylor' show poor quality on the first iterations, but still manage to converge faster, than \verb'svi-AdaDelta'. On both datasets the \verb'vi-JJ-hybrid' method works similar to \verb'vi-JJ' and \verb'vi-Taylor', but shows better quality on first iterations on the {\it magic telescope} data. \verb'vi-JJ-full' can't converge to a reasonable quality on both datasets.

		Fig.~\ref{big} provides the results on big {\it cod-rna} and {\it skin\_nonskin} datasets. On these datasets the \verb'vi-JJ-full' once again fails to achieve a reasonable quality, while the other methods work similarly.

\begin{figure}[!t]
			\center
			\scalebox{0.85}{
			\includegraphics{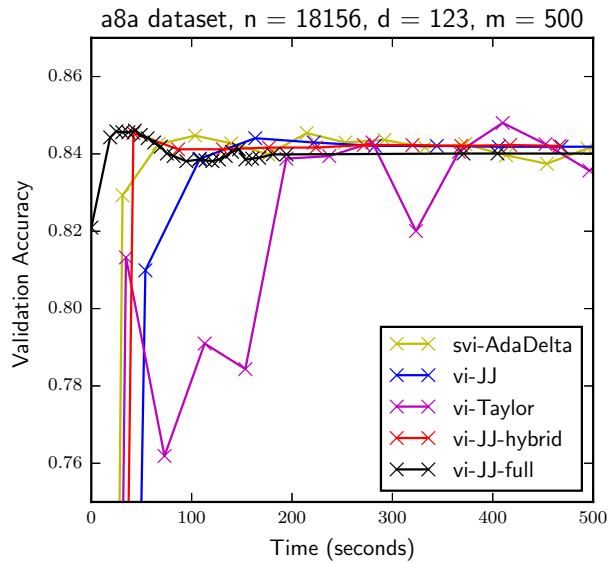}}
			\caption{Methods performance on the a8a dataset}
			\label{big_d}
		\end{figure}

		Finally, the results on {\it a8a} data are provided in fig.~\ref{big_d}. Here we use a rather big amount of $m = 500$ inducing inputs. As we can see, the \verb'vi-JJ-full' and \verb'vi-JJ-hybrid' are the fastest to achieve the optimal quality. The \verb'svi-AdaDelta' method also converges reasonably fast, while the \verb'vi-JJ' and \verb'vi-Taylor' struggle a little bit.

		In general \verb'vi-JJ', \verb'vi-Taylor' and \verb'vi-JJ-hybrid' methods perform similar to \verb'svi-AdaDelta' method. On the big dataset {\it skin\_nonskin} with only $3$ features the \verb'vi-JJ-hybrid' is a little bit slower than the stochastic \verb'svi-AdaDelta', but on all the other datasets it is better. The \verb'vi-Taylor' and \verb'vi-JJ' struggle with {\it a8a}, but are otherwise comparable to \verb'vi-JJ-hybrid'. The stochastic \verb'svi-AdaDelta' method performs poorly on small datasets and even on the big {\it skin\_nonskin} data doesn't manage to substantially outperform the other methods, even provided a good value of learning rate. Finally, \verb'vi-JJ-full' works well on small data and on the {\it a8a}, but on all the other datasets it doesn't manage to achieve a reasonable quality.

\section{Conclusion}
	In this paper we presented a new approach to training variational inducing input Gaussian process classification. We derived two new tractable evidence lower bounds and described several ways to maximize them. The resulting methods \verb'vi-JJ', \verb'vi-JJ-full', \verb'vi-JJ-hybrid' and \verb'vi-Taylor' are similar to the method of~\cite{Titsias} for GP-regression.

We provided an experimental comparison of our methods with the current state-of-the-art method \verb'svi-AdaDelta' of~\cite{SviClass}. In our experimental setting, our approach proved to be more practical, as it converges to the optimal quality as fast as the \verb'svi-AdaDelta' method without requiring the user to manually choose the parameters of the optimization method.

The four described \verb'vi' methods showed similar performance and it's hard to distinguish them. However, note that the \verb'vi-Taylor' approach is more general and can be applied to the likelihood functions, that are not logistic. We could also easily derive a method, similar to \verb'vi-JJ-hybrid' and \verb'vi-JJ-full' for the non-logistic case, but it is out of scope of this paper.

\end{document}